\pdfoutput=1
\documentclass{article}
\usepackage{ijcai20}

\usepackage{times}
\usepackage{soul}
\usepackage{url}
\usepackage[hidelinks]{hyperref}
\usepackage[utf8]{inputenc}
\usepackage[small]{caption}
\usepackage{graphicx}
\usepackage{amsmath}
\usepackage{amsthm}
\usepackage{booktabs}
\usepackage{algorithm}
\usepackage{algorithmic}
\usepackage{bm}
\usepackage{array}
\newcommand{\PreserveBackslash}[1]{\let\temp=\\#1\let\\=\temp}
\newcolumntype{C}[1]{>{\PreserveBackslash\centering}p{#1}}
\newcolumntype{R}[1]{>{\PreserveBackslash\raggedleft}p{#1}}
\newcolumntype{L}[1]{>{\PreserveBackslash\raggedright}p{#1}}
\usepackage{multirow}

\newtheorem{example}{Example}
\newtheorem{theorem}{Theorem}

\title{DAM: Deliberation, Abandon and Memory Networks for Generating Detailed and Non-repetitive Responses in Visual Dialogue}

\author{
Xiaoze Jiang$^{1,2}$\footnote{Equal contribution. This
work is done when Xiaoze Jiang is an intern in IIE, CAS.}\and
Jing Yu$^{1,3*}$\footnote{Corresponding author.}\and
Yajing Sun$^{1,3}$\and\\
Zengchang Qin$^{2,4\dagger}$\and 
Zihao Zhu$^{1,3}$\and
Yue Hu$^{1,3}$\And
Qi Wu$^{5}$ \\
\affiliations
$^1$Institute of Information Engineering, Chinese Academy of Sciences, Beijing, China\\
$^2$Intelligent Computing and Machine Learning Lab, School of ASEE, Beihang University, Beijing, China\\
$^3$School of Cyber Security, University of Chinese Academy of Sciences, Beijing, China\\
$^4$AI Research, Codemao Inc. 
$^5$University of Adelaide, Australia\\
\emails 
\{yujing02, sunyajing, zhuzihao, huyue\}@iie.ac.cn,
\{xzjiang, zcqin\}@buaa.edu.cn,
qi.wu01@adelaide.edu.au
}

\begin{document}

\maketitle

\begin{abstract}
Visual Dialogue task requires an agent to be engaged in a conversation with human about an image. The ability of generating detailed and non-repetitive responses is crucial for the agent to achieve human-like conversation. In this paper, we propose a novel generative decoding architecture to generate high-quality responses, which moves away from decoding the whole encoded semantics towards the design that advocates both transparency and flexibility. In this architecture, word generation is decomposed into a series of attention-based information selection steps, performed by the novel recurrent Deliberation, Abandon and Memory (DAM) module. Each DAM module performs an adaptive combination of the response-level semantics captured from the encoder and the word-level semantics specifically selected for generating each word. Therefore, the responses contain more detailed and non-repetitive descriptions while maintaining the semantic accuracy. Furthermore, DAM is flexible to cooperate with existing visual dialogue encoders and adaptive to the encoder structures by constraining the information selection mode in DAM. We apply DAM to three typical encoders and verify the performance on the VisDial v1.0 dataset. Experimental results show that the proposed models achieve new state-of-the-art performance with high-quality responses. The code is available at https://github.com/JXZe/DAM.

\end{abstract}

\section{Introduction}

Visual Dialogue \cite{Das2017Visual} is a task that requires an agent to answer a series of questions grounded in an image, demanding the agent to reason about both visual content and dialogue history. There are two kinds of typical approaches to this task  \cite{Das2017Visual}: \textit{discriminative} and \textit{generative}. Discriminative approach learns to select the best response in a candidate list, while generative approach may generate new responses that are not provided in the pre-constructed repository. The discriminative approach is relatively easier since the grammaticality and accuracy are guaranteed in the human-written responses.  However, the retrieved responses are limited by the capacity of the pre-constructed repository. Even the best matched response may not be exactly appropriate since most cases are not tailored for the on-going questions \cite{qi2019two}. Therefore, the generative ability is crucial to achieve human-like conversation by synthesizing more factual and flexible responses accordingly. 
\begin{figure}[t]
\setlength{\belowcaptionskip}{-4mm} 
\setlength{\abovecaptionskip}{5pt}
\centering
\includegraphics[width=7.9cm]{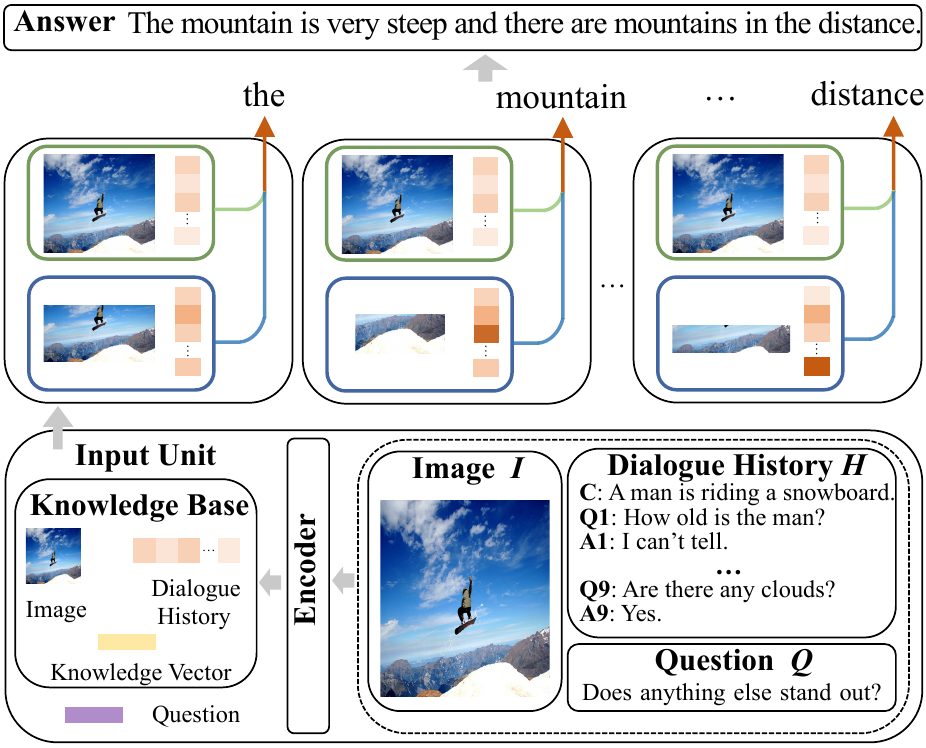}
\caption{An illustration of DAM. The encoder encodes visual dialogue information into Knowledge Base (KB). DAM adaptively composites the information from response-level (green block) and word-level (blue block) to generate word at each decoding step.}

\label{basicIdea}
\end{figure}
The typical solution for the generative visual dialogue system is based on the encoder-decoder framework \cite{yang2019making}. The encoder aims to capture the semantics of the image, question and dialogue history by embeddings, while the decoder decodes these embeddings to a response by recurrent neural networks (RNN) \cite{HopfieldNeural}. Due to the difficulty of generation, the majority of previous works \cite{Niu2018Recursive} have focused on designing more comprehensive encoder structures to make use of different aspects of information from the input. Though these methods achieve promising improvement, they still have obvious limitations, such as generating inaccurate details and repetitive words or phrases.

To tackle the above problems, we propose to adaptively incorporate more detailed information from the encoder for generating each word in the decoding process. Specifically, we propose a recurrent \textbf{Deliberation, Abandon} and \textbf{Memory} (\textbf{DAM}) module, a novel architecture of generative decoder to address the above two issues. As shown in Figure \ref{basicIdea}, on the one hand, DAM incorporates the global information in the response-level to keep semantic coherence. On the other hand, DAM pays attention to capture the related and unique details in the word-level by designing Deliberation Unit guided by the current generated word. To further reduce repetition, we devise Abandon Unit to select the unique information for the current word. In the end, Memory Unit integrates the derived word-level and response-level semantics into the memory state for word generation, which contributes to the unification of semantic coherence and the richness of details. With recurrent connections between the DAM cells inspired by LSTM \cite{hochreiter1997long}, the network is capable of generating visual-grounded details in a progressive manner and remarkably eliminates repetition by coverage control. Note that DAM is a universal architecture that can be combined with existing visual dialogue models by adapting the Deliberation Unit to the corresponding encoder. To show the effectiveness of DAM, we propose three models by combining DAM with three typical visual dialogue encoders, including Late Fusion encoder \cite{Das2017Visual} for general feature fusion, Memory Network encoder \cite{Das2017Visual} for dialogue history reasoning, and DualVD encoder \cite{jiang2019daulvd} for visual-semantic image understanding. We show that the performance of baseline models is consistently improved by combining with DAM.

The main contributions are summarized as follows: 
(1) We propose a novel generative decoder DAM to generate more detailed and less repetitive responses. DAM contains a compositive structure that leverages the complementary information from both response-level and word-level, which guarantees the accuracy and richness of the responses. (2) DAM is universal to cooperate with existing visual dialogue encoders by constraining the information selection mode to adapt to different encoder structures. (3) We demonstrate the module's capability, generality and interpretability on the VisDial v1.0 dataset. DAM consistently improves the performance of existing models and achieves a new state-of-the-art 60.93\% on NDCG for the generative task.

\section{Related Work}

\paragraph{Visual Dialogue.} 
  Most previous works focused on discriminative approaches \cite{Zheng2019Reasoning,schwartz2019factor,Kottur2018Visual,kong2018visual}  and achieved great progress. However, generative approaches, which are more practical in realistic applications, typically perform  inferior to the discriminative approaches. \cite{wu2018areyou} combined reinforcement learning with generative adversarial networks \cite{goodfellow2014generative} to generate human-like answers. \cite{zhang2019generative} introduced negative responses to generative model to reduce safe responses. \cite{chen2019dmrm} proposed a multi-hop reasoning model to generate more accurate responses. However, how to generate less repetitive and more detailed responses has been less studied. Our work devotes to reducing the repetition and improving the richness in responses via designing a universal generative decoder by selecting more related information for generating the current word from response-level and word-level semantics.

\paragraph{Generation-based Dialogue Systems.} The typical solution adopts the sequence-to-sequence (seq2seq) framework \cite{madotto-etal-2018-mem2seq,xu-etal-2019-neural} and uses RNN to generate responses. Existing works studied diverse aspects of generation, including expressing specific emotions \cite{song-etal-2019-generating,rashkin2019towards}, introducing new topics \cite{xu-etal-2019-neural}, generating robust task-oriented responses \cite{peng-etal-2018-deep,lei2018sequicity}, improving the richness  \cite{tian-etal-2019-learning} and reducing repetition  \cite{shao2017generating}, {\it etc}. \cite{see2017get} assigned {\it pointing} to copy words from the source text to improve the richness of sentences and used {\it coverage mechanism} to reduce repetition. The problem of reducing repetition of response has been less studied in visual dialogue. What's more, the methods in visual dialogue cannot adopt {\it pointing} to copy words directly, since the {\it pointing} clues come from image and dialogue history in visual dialogue. One limitation of coverage mechanism is that it reduces repetition by rigid constraints of the loss function, which may result in the missing of essential words. Intuitively, understanding the input information comprehensively and capturing word-specific semantics can also reduce repetition. Inspired by this intuition, we propose a novel visual dialogue decoder to generate less repetitive and more detailed responses by considering the encoder structure and adaptively selecting and decoding information from the encoder.

\section{Methodology}

The visual dialogue task can be described as follows: given an image $I$ and its caption $C$, a dialogue history till round $t$-$1$, $H_t = \left\{ C, (Q_1,A_1),...,(Q_{t-1},A_{t-1})\right\}$, and the current question $Q_t$, the task aims to generate an accurate response $A_t$. Our work mainly focuses on the design of a novel generative decoder architecture DAM. To prove the effectiveness of DAM, we combine it with three typical encoders: Late Fusion (LF), Memory Network (MN) and the state-of-the-art Dual-coding Visual Dialogue (DualVD). In this section, we will introduce (1) the typical encoder-decoder generative model in visual dialogue, (2) the structure of our proposed generative decoder, and (3) the combination strategies of our decoder with the three typical encoders.




\subsection{Encoder-Decoder Generative Model}
\label{base-ender}
Our proposed DAM network is an advancement of the typical generative decoder with deliberation and control abilities. In this section, we first introduce the typical generative visual dialogue encoder-decoder model. Encoder encodes the image $I$, dialogue history $H_t$ and current question $Q_t$ by a hidden state called knowledge vector $K_t$ (for conciseness, $t$ is omitted below). On each decoding step $\tau$, the decoder, typically using a single-layer unbidirectional LSTM, receives the word embedding of previous generated word $x_{\tau-1}$ and previous hidden state $s_{\tau-1}$ (the output knowledge vector $K$ from encoder serves as the initial hidden state) and outputs a decoded vector $a_\tau$. Then the probability distribution $P_\tau$ over the vocabulary can be computed by:
\begin{align}
P_{\tau} =    softmax( \bm{w}_p^T a_\tau+b_p)
\label{p_tau}
\end{align}
The word with the highest probability is selected as the predicted word and the model is trained by log-likelihood loss. 

\subsection{The DAM Decoder} 





\begin{figure}[t]
\setlength{\abovecaptionskip}{5pt}
\setlength{\belowcaptionskip}{-3mm} 
\centering
\includegraphics[width=7.5cm]{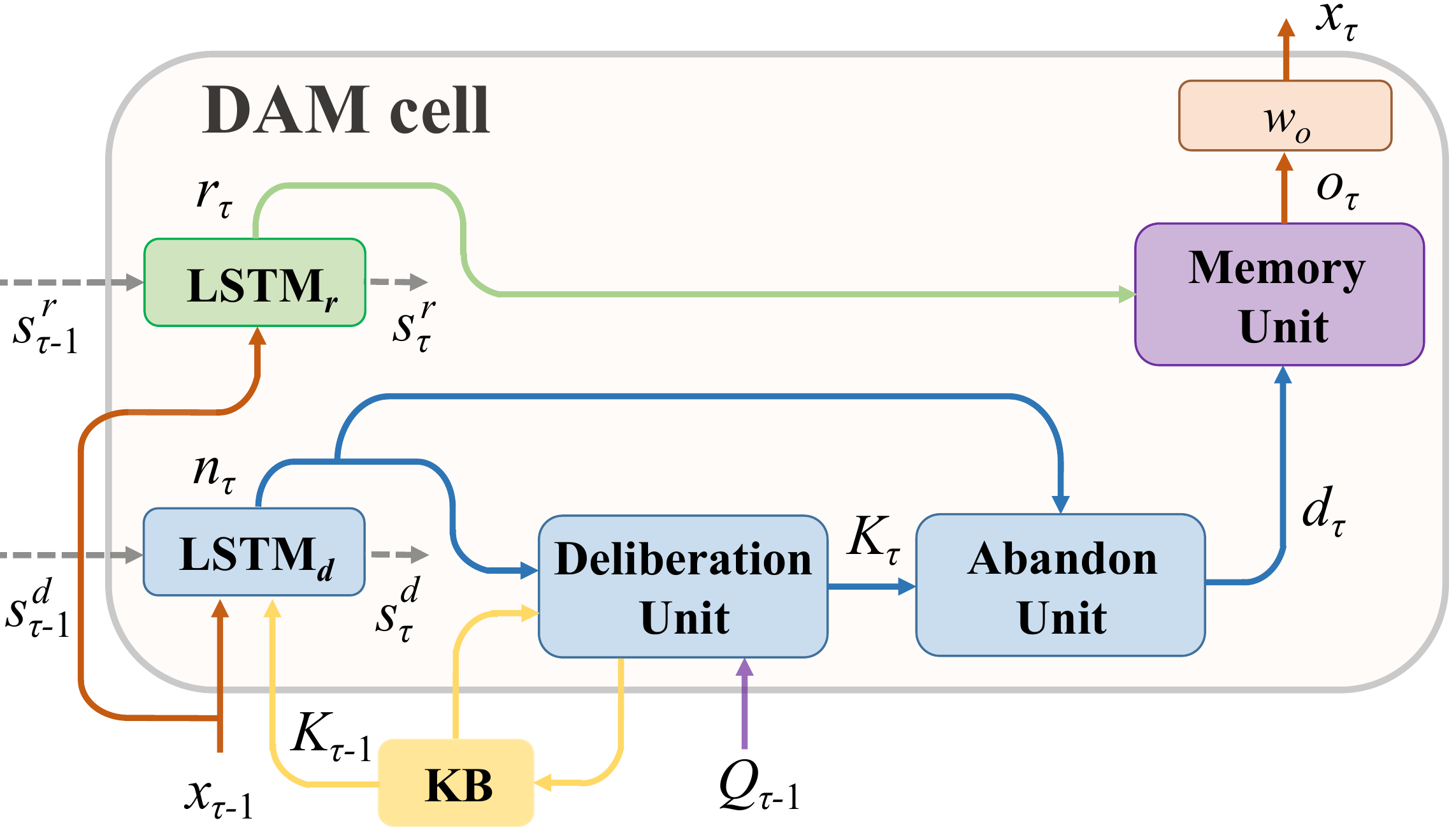}
\caption{Overview structure of DAM. It consists of RSL (green part), WDL (blue part) and information fusion module (purple part). For the $\tau$ generation step, the inputs of DAM contain the question embedding $Q_{\tau-1}$, the knowledge embedding $K_{\tau-1}$, the previous generated word embedding $x_{\tau-1}$ and the previous hidden states $s_{\tau-1}^{r}$ and $s_{\tau-1}^{d}$.}

\label{model_pic}
\end{figure}

DAM is a novel compositive decoder that can be incorporated with standard sequence-to-sequence generation framework. It helps to improve the richness of semantic details as well as discouraging repetition in the responses. As shown in Figure \ref{model_pic}, DAM consists of response-level semantic decode layer (RSL), word-level detail decode layer (WDL) and information fusion module (Memory Unit). 
RSL is responsible for capturing the global information to guarantee the response's fluency and correctness. However, the global information lacks the detailed semantics, for the current word and the rigid-decoding mode in LSTM tends to generate repeated words. WDL incorporates the essential and unique visual dialogue contents (i.e.  question,  dialogue history and image) into the generation of current word to enrich the word-level details. The structure of WDL consists of an LSTM, Deliberation Unit and Abandon Unit. Finally, Memory Unit is responsible for adaptively fusing both the response-level and word-level information.

\setlength{\parskip}{5pt}
\noindent\textbf{Response-Level Semantic Decode Layer (RSL)}
\setlength{\parskip}{0pt}

\noindent When answering a question about an image, human needs to capture the global semantic information to decide the main ideas and content for the responses. In our model, we regard the embedded information from the encoder as global semantic information, and denote it as knowledge vector $K$. $K$ is used for providing the response-level semantics in the generation process. The response-level information $r_\tau$ for generating the current word is computed as:
\begin{align}
\label{lstmm}  r_\tau &=LSTM_r(x_{\tau-1},s_{\tau-1}^{r})
\end{align}
where $x_{\tau-1}$ is the previous generated word and $s_{\tau-1}^{r}$ is the memory state of LSTM$_r$.

\setlength{\parskip}{5pt}
\noindent\textbf{Word-Level Detail Decode Layer (WDL)}
\setlength{\parskip}{0pt}

\noindent On the one hand, the response-level information lacks the details of the image and dialogue history, providing rigid clues for generating different words. On the other hand, response-level information changes slightly with the recurrent word generation process and results in repetitive words or phrases. To solve these problems, it's critical to enrich the decoding vector with more detailed question-relevant information that is unique for current generated word.

For generating the $\tau^{th}$ word, we first adaptively capture word-relevant information from the encoded knowledge information along with previous generated word and previous hidden state via LSTM$_d$: 
\begin{align}
\label{lstmd}  n_\tau &=LSTM_d([x_{\tau-1}, K_{\tau-1}],s_{\tau-1}^{d})
\end{align} 
where  ``$[\cdot ,\cdot ]$'' denotes concatenation, $K_{\tau-1}$ is the updated knowledge vector in the $\tau$-$1$ step and $s_{\tau-1}^{d}$ is the memory state of LSTM$_d$. Since $n_\tau$ only capture the global semantics from the encoder, we further incorporate the structure-adaptive local semantics from the encoder via the Deliberation Unit. Finally, we propose the Abandon Unit to filter out the redundant information while enhancing the word-specific information from both global and local clues. The Deliberation Unit and the Abandon Unit are detailed below.


\paragraph{Deliberation Unit.} It aims to adaptively leverage the encoder structure to extract the most related and detailed information for current word generation. Specifically, we first capture the significant information in the question under the guidance of the global semantic vector $n_\tau$. Guided by the upgraded question representation, we adopt structure-adaptive strategies to different encoder structures to select image and dialogue history information. In the end, we get the detailed question-related information by fusing the information of question, dialogue history and image. Compared with most existing decoders that merely use the encoded embedding without considering the diverse encoder structures, our proposed Deliberation Unit provides a flexible strategy to derive more detailed information by taking the advantages of the elaborate encoders. To prove the effectiveness of DAM, we combine it with three typical encoders, including LF encoder for the general feature fusion, MN encoder for dialogue history reasoning and DualVD encoder for visual-semantic image understanding. The details of Deliberation Unit adaptive to these three encoders will be introduced in Section \ref{var-de-u}.

\begin{figure*}[t]
\setlength{\abovecaptionskip}{5pt}
\setlength{\belowcaptionskip}{-4mm} 
\centering
\includegraphics[width=17.5cm]{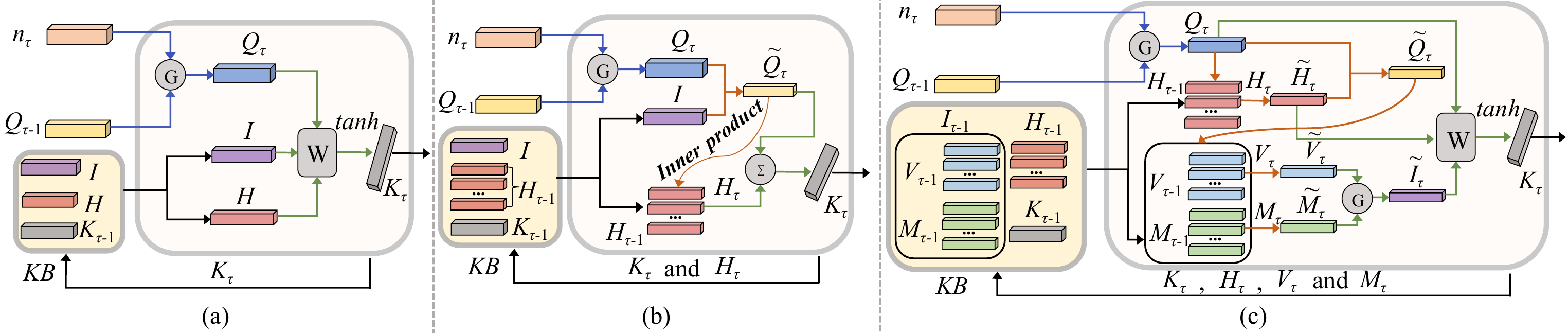}
\caption{The illustration of Deliberation Unit adaptive to LF Encoder (a), MN Encoder (b) and DualVD Encoder (c), where ``G": gate operation, $V$: visual representation of image, $M$ semantic representation of image. Different colored lines represent different update steps: {\it blue lines}: word-guided question information update, {\it orange lines}: question-guided information update, {\it green lines}: general feature fusion.}
\label{de-module}
\end{figure*}


\paragraph{Abandon Unit.} It further filters out the redundant information while enhancing the word-specific information from both the global and local encoded clues. Specifically, Abandon Unit updates current generated decoding vector $n_\tau$ by combining detailed knowledge information $K_\tau$ with $n_\tau$ via a gate operation and achieves the final word-level embedding $d_\tau$:
\begin{align}
gate_{\tau}^a =  \sigma (\bm{W}_a [n_{\tau}, K_{\tau}] + b_a)&\\
d_\tau =   gate_{\tau}^a \circ  [n_{\tau}, K_{\tau}]&
\label{gate_a} 
\end{align}

\noindent where ``$\circ$'' denotes the element-wise product. 

\setlength{\parskip}{5pt}
\noindent\textbf{Two Level Information Fusion}
\setlength{\parskip}{0pt}

\noindent The information from RSL and WDL is complementary to each other. We design {\it Memory Unit} to combine the two kinds of information for word prediction. Memory Unit selects response-level information to control the global semantics in response and tracks the word-level information for generating more detailed and less repeated response via a gate operation:
\begin{align}
\label{gate_m} gate_{\tau}^m =  \sigma (\bm{W}_m [r_{\tau}, d_{\tau}] + b_m)&\\
o_\tau = gate_{\tau}^m \circ  [r_{\tau}, d_{\tau}]&
\end{align}
The generated word $x_\tau$ with the maximum value in the probability distribution $P_\tau^o$ is selected as the predicted word. $P_\tau^o$ is computed as:
\begin{align}
P_{\tau}^o = softmax( \bm{w}_o^T o_\tau+b_o)
\label{new}
\end{align}

\subsection{Variants of Deliberation Unit}
\label{var-de-u}

Guided by the question and current generated word state $n_\tau$, Deliberation Unit captures more detailed information from encoder-specific structures. The Deliberation Unit mainly contains three steps: (1) word-guided question information update,  (2)  question-guided information update, and (3) general feature fusion. The last two steps are adaptive to different encoders while the first step keeps unchanged. To select the most related information for current generated word, we first update question information $Q_{\tau-1}$ with $n_\tau$:
\begin{align}
\label{gate_q} gate_{\tau}^q =  \sigma (\bm{W}_q [Q_{\tau-1}, n_{\tau}] + b_q)&\\
Q_{\tau} = \bm{W}_1 ( gate_{\tau}^q  \circ [Q_{\tau-1}, n_{\tau}] )&
\label{gate_q}
\end{align}
We will introduce the next two steps adaptive to LF, MN and DualVD encoders below. It should be noted that the parameters of the  Deliberation Unit are independent of its encoder.

\paragraph{Deliberation Unit Adaptive to LF Encoder.} LF Encoder focuses on multi-modal information fusion without complex information reasoning. In our decoder, we merely fuse the updated question information $Q_\tau$ with dialogue history $H$ and image $I$ from the encoder without question-guided information update step as shown in Figure \ref{de-module}(a).


\paragraph{Deliberation Unit Adaptive to MN Encoder.}  MN Encoder focuses on the dialogue history reasoning. Compared with Deliberation Unit for LF Encoder, we further add question-guided information update step to reason over dialogue history via attention mechanism before general feature fusion as shown in Figure \ref{de-module}(b).


\paragraph{Deliberation Unit Adaptive to DualVD Encoder.}
 DualVD Encoder focuses on the visual-semantic image understanding. As shown in Figure \ref{de-module}(c), for the question-guided information update step, we first concatenate updated question and dialogue history to form the query vector $\widetilde{Q}_{\tau}$, and assign $\widetilde{Q}_{\tau}$ to guide the update of image from visual and semantic aspects respectively. For the feature fusion step, we utilize the gate operation between visual and semantic image representation ($\widetilde{V}_{\tau}$ and $\widetilde{T}_{\tau}$) to obtain the updated image representation.

\section{Experiments}
\label{sec:experiments}


\paragraph{Dataset.} We conduct extensive experiments on VisDial v1.0 dataset \cite{Das2017Visual} constructed based on MSCOCO images and captions. VisDial v1.0 is split into training, validation and test sets. The training set consists of dialogues on 120k images from COCO-trainval while the validation and test sets are consisting of dialogues on an additional 10k COCO-like images from Flickr.  

\paragraph{Evaluation Metrics.} Following \cite{Das2017Visual}, we rank the 100 candidate answers based on their posterior probabilities and evaluate the performance by retrieval metrics: mean reciprocal rank (MRR), recall@$k$ ($k$ =1, 5, 10), mean rank of human response (Mean) and normalized discounted cumulative gain (NDCG). Lower value for Mean and higher value for other metrics are desired.

\begin{table}[t] 
\setlength{\abovecaptionskip}{4pt}
\setlength{\belowcaptionskip}{-4mm} 
\centering
\resizebox{.95\columnwidth}{!}{
\begin{tabular} {L{4.3cm}C{0.6cm}C{0.6cm}C{0.6cm}C{0.69cm}C{0.6cm}C{0.77cm}}
\hline                       
Model & MRR & R\textsl{@}1 & R\textsl{@}5 & R\textsl{@}10 & Mean & NDCG  \\
\hline  
HCIAE-G \cite{lu2017best} &49.07 & 39.72 & 58.23 & 64.73 &  18.43 & 59.70 \\
CoAtt-G \cite{wu2018areyou} & 49.64 & 40.09 & 59.37 & 65.92 & 17.86 & 59.24 \\ 
Primary-G \cite{guo2019image} & 49.01&38.54&59.82&66.94&16.69&-\\
ReDAN-G \cite{Gan2019Multi}& 49.60 & 39.95&  59.32 & 65.97 &  17.79  & 59.41 \\
DMRM \cite{chen2019dmrm} &50.16&40.15&60.02&67.21&\textbf{15.19}&-\\
\hline
LF-G \cite{Das2017Visual}&44.67&34.84&53.64&59.69&21.11&52.23\\
MN-G \cite{Das2017Visual}&45.51&35.40&54.91&61.20&20.24&51.86\\
DualVD-G \cite{jiang2019daulvd} & 49.78 & 39.96 & 59.96 & 66.62 & 17.49 & 60.08\\
\hline
\hline 
\textbf{LF-DAM (ours)}&45.08&35.01&54.48&60.57&20.83&52.68\\
\textbf{MN-DAM (ours)} &46.16&35.87&55.99&62.45&19.57&52.82\\
\textbf{DualVD-DAM (ours)} &\textbf{50.51}&\textbf{40.53}&\textbf{60.84}&\textbf{67.94}&16.65&\textbf{60.93 }\\
\hline  
\end{tabular}  
}
\caption{Result comparison on validation set of VisDial v1.0.}
\label{v1}
\end{table}

\paragraph{Implementation Details.} To build the vocabulary, we retain words in the dataset with word frequency greater than 5. The vocabulary contains 10366 words. The hidden states and cell states of LSTM$_d$ are randomly initialized while LSTM$_r$ is using the output knowledge vector $K$ from encoder as the initial hidden state and randomly initializing cell state. The maximum sentence length of the responses is set to 20. The hidden state size of all the LSTM blocks is set to 512 and the dimension of each gate is set to 1024. The Adam optimizer \cite{Kingma2014Adam} is used with the initial learning rate of 1e-3 and final learning rate of 3.4e-4 via cosine annealing strategy with 16 epochs. The batch size is set to 15.

\subsection{State-of-the-Art Comparison}

As shown in Table \ref{v1}, we compare our models (third block) with SOTA generative models (first block) and baseline models (second block, re-trained by us). ReDAN-G and DMRM adopted complex multi-step reasoning, while HCIAE-G, CoAtt-G and Primary-G are attention-based models. For fairness, we only compare the original generative ability without re-ranking. We just replace the decoders in baseline models by our proposed DAM. Compared with the baseline models, our models outperform them on all the metrics, which indicates the complementary advantages between DAM and existing encoders in visual dialogue. Though DualVD-G performs lower than DMRM on {\it Mean}, DualVD-DAM outperforms DMRM on all the other metrics without multi-step reasoning, which is the advantages in DMRM over our models.

\subsection{Ablation Study}
\label{ab-s-study-sec}

\setlength{\parskip}{5pt}
\noindent\textbf{The  Effectiveness of Each Unit}
\setlength{\parskip}{0pt}

\begin{table}[t] 
\setlength{\abovecaptionskip}{5pt}
\setlength{\belowcaptionskip}{-3mm}
\centering
\resizebox{.9\columnwidth}{!}{
\begin{tabular} {L{2cm}L{1.5cm}C{0.6cm}C{0.6cm}C{0.6cm}C{0.69cm}C{0.6cm}C{0.77cm}}

\hline                       
Base Model & Model & MRR & R\textsl{@}1 & R\textsl{@}5 & R\textsl{@}10 & Mean & NDCG  \\
\hline
\hline  
\multirow{4}{*}{LF-DAM}& 2LSTM & 44.43 &34.53&53.55&59.48&21.38&51.99\\
&2L-M&44.77&34.85&54.06&60.03&21.13&52.04\\
&2L-DM&45.06 & 34.90 & 54.24 & 60.39 & 20.87 & 52.58\\
&2L-DAM&\textbf{45.08}&\textbf{35.01}&\textbf{54.48}&\textbf{60.57}&\textbf{20.83}&\textbf{52.68}\\

\hline  
\hline
\multirow{4}{*}{MN-DAM} &2LSTM&45.58&35.27&55.38&61.54&19.96&52.38\\
&2L-M&45.67 &35.29&55.57&61.97&19.91&52.11\\
&2L-DM&45.77 & 35.53&  55.40 & 62.05 &  19.95  & 52.51\\
&2L-DAM&\textbf{46.16}&\textbf{35.87}&\textbf{55.99}&\textbf{62.45}&\textbf{19.57}&\textbf{52.82}\\

\hline  
\hline
\multirow{4}{*}{DualVD-DAM} & 2LSTM&49.72&40.04&59.52&66.41&17.62&59.79\\
&2L-M&50.09&40.38&59.94&66.77&17.31&59.85\\
&2L-DM&50.20&40.33&60.22&67.48&17.15&59.72\\
&2L-DAM&\textbf{50.51}&\textbf{40.53}&\textbf{60.84}&\textbf{67.94}&\textbf{16.65}&\textbf{60.93 }\\

\hline  
\end{tabular}  
}
\caption{Ablation study of each unit on VisDial v1.0 validation set.}
\label{abf}
\end{table}

\noindent We consider the following ablation models to illustrate the effectiveness of each unit of our model:
1) \textbf{2L-DAM}: this is our full model that adaptively selects related information for decoding. 2) \textbf{2L-DM}: full model w/o Abandon Unit. 3) \textbf{2L-M}: 2L-DM w/o Deliberation Unit. 4) \textbf{2-LSTM}: 2L-M w/o Memory Unit.
As shown in Table \ref{abf}, taking DualVD-DAM for example, the  MRR values increase by  0.37\%, 0.11\% and 0.31\% respectively when introducing the Memory Unit (2L-M), Deliberation Unit (2L-DM) and Abandon Unit (2L-DAM) to the baseline model (2-LSTM) progressively.  Similar trend exists in LF-DAM and MN-DAM, which indicates the effectiveness of each unit in DAM. Since the space limitation and similar observations, we show the ablation studies on DualVD-DAM in the following experiments.

\setlength{\parskip}{5pt}
\noindent\textbf{The  Effectiveness of Two-Level Decode Structure}
\setlength{\parskip}{0pt}

\begin{table}[t] 
\setlength{\abovecaptionskip}{5pt}
\setlength{\belowcaptionskip}{-3mm} 
\centering
\resizebox{.9\columnwidth}{!}{
\begin{tabular}{L{4cm}C{1cm}C{1cm}C{1.5cm}C{1.8cm}}
\hline                
Model &  M1 $\uparrow$ & M2 $\uparrow$ & Repetition$\downarrow$ & Richness $\uparrow$ \\
\hline
RSL(DualVD-G): RSL only &0.60 &0.47&0.20&0.03\\
WDL: WDL only & 0.69 & 0.54 &0.07&\textbf{0.15}\\
\textbf{DualVD-DAM} &\textbf{0.75} & \textbf{0.61} &\textbf{0.01}& 0.13\\
\hline
\end{tabular}  
}
\caption{Human evaluation of 100 sample responses on VisDial v1.0 validation set. M1: percentage of responses that pass the Turing Test. M2: percentage of responses that are evaluated better or equal to human responses. Repetition: percentage of responses that have meaningless repeated words. Richness: percentage of responses that contain detailed content to answer the question.}
\label{human}
\end{table}

\begin{figure}[t]
\setlength{\abovecaptionskip}{5pt}
\setlength{\belowcaptionskip}{-4mm} 
\centering
\includegraphics[width=8.5cm]{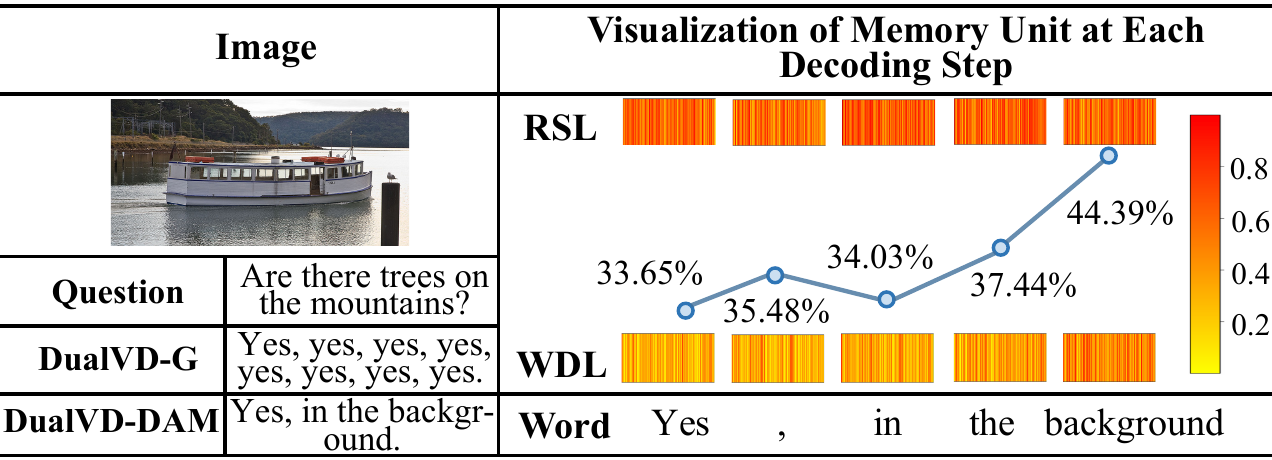}

\caption{Visualization of gate values in the Memory Unit. Yellow thermogram: gate values of RSL and WDL computed in Eq. (\ref{gate_m}). Blue polyline: ratio of total gate values of WDL.}
\label{gate_v}
\end{figure}

\noindent To prove the complementary advantages of the response-level semantic decode layer (RSL) and the word-level detail decode layer (WDL), and to figure out the information selection mode, we first conduct \textbf{Human Study} on the effectiveness of RSL, WDL  and the full model DualVD-DAM, and then visualize the gate values of Memory Unit to reveal the information selection mode.


\paragraph{Complementary Advantages.} In the human study, we follow \cite{wu2018areyou} to sample 100 results from VisDial v1.0 validation set and ask 3 persons to evaluate the quality of the last response in the dialogue. Distinct from previous works, we add \emph{Repetition} and \emph{Richness} metrics, and for all metrics, we record the score when at least 2 persons agree. As shown in Table \ref{human}, WDL performs best on {\it Richness} and reduces the {\it Repetition} by 0.13 compared to RSL, which indicates that WDL contributes to the increase of detailed information and the decrease of repetition in the response. After incorporating RSL and Memory Unit with WDL, the {\it repetition} further reduces by 0.06 while M1 and M2 outperform by 0.06 and 0.07 respectively, which proves the complementary advantages between these two level information. We also notice that {\it Richness} decreases slightly. This is mainly because the information from RSL concentrates more attention on the global information, rather than detailed information.
 
\paragraph{Information Selection Mode.} We  visualize the gate values in the Memory Unit for DualVD-DAM to demonstrate the  information selection mode of the two level information. As shown in Figure \ref{gate_v}, we can observe that the ratio of gate values for RSL is always higher than that for WDL. It indicates that the response-level information in RSL plays the predominant role in guiding the response generation. Another obvious phenomenon is that the ratio of gate values for WDL increases rapidly when generating the last word, which can be viewed as a  signal to stop the response generation in time when the response already covers the complete semantics. It may due to the fact that WDL captures word-level information and is sensitive to the word repetition, which is beneficial to avoid repetitive word generation.

\begin{figure*}[t]
\setlength{\abovecaptionskip}{5pt}
\setlength{\belowcaptionskip}{-3mm} 
\centering
\includegraphics[width=17cm]{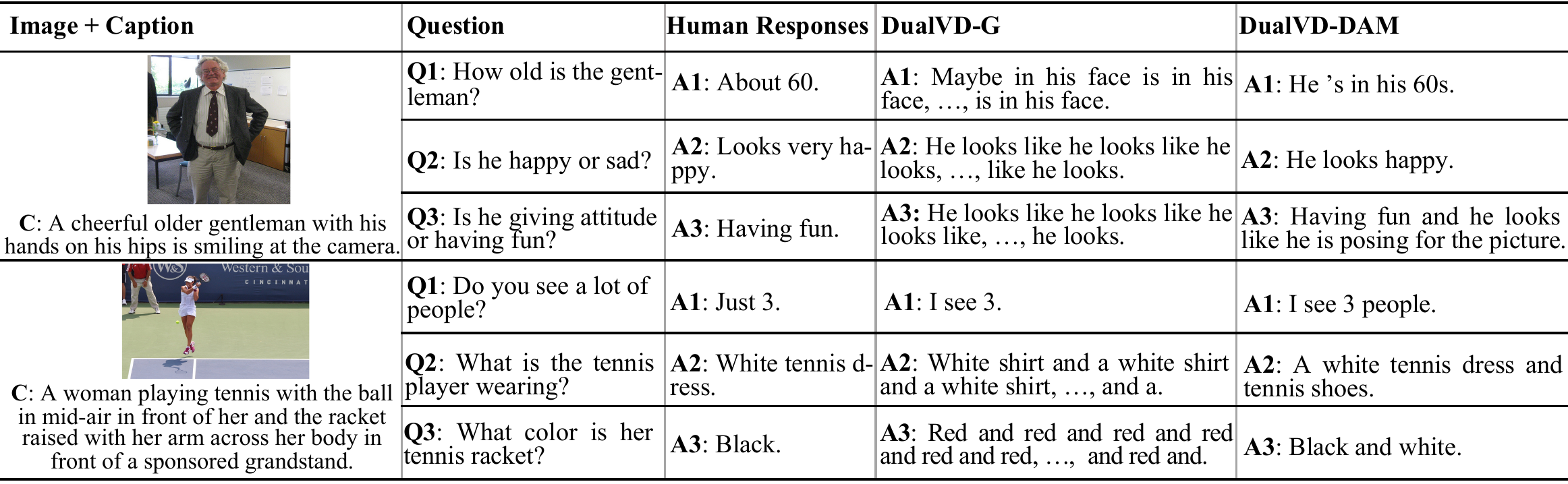}
\caption{Qualitative results of DualVD-DAM comparing to Human Responses and DualVD-G, where ``\emph{...}" are omitted repeated words.}
\label{visual-se}
\end{figure*}

\begin{figure*}[t]
\setlength{\abovecaptionskip}{5pt}
\setlength{\belowcaptionskip}{-4mm} 
\centering
\includegraphics[width=17.5cm]{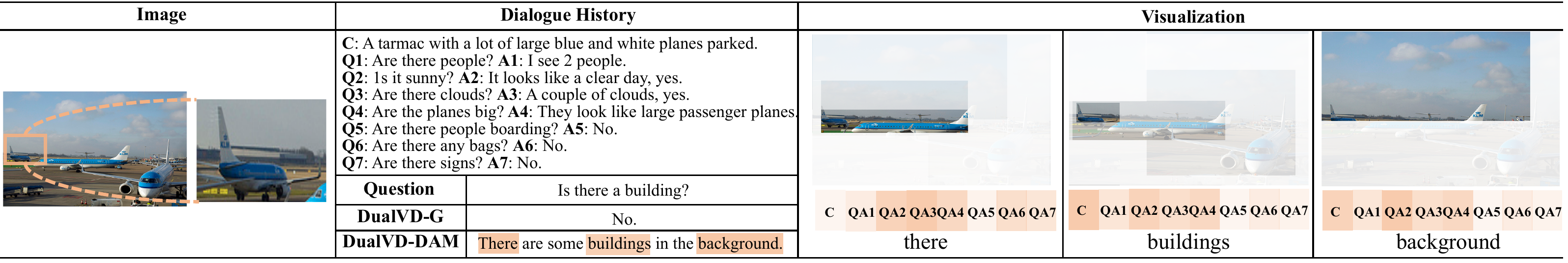}
\caption{Visualization of the evidence when generating the response by DualVD-DAM. The essential visual regions and dialogue history for answering the question are highlighted in the last three columns. The attention weights of visual regions and dialogue history are visualized, where clearer region and darker orange color indicates higher attention weight.}
\label{time}
\end{figure*}

\setlength{\parskip}{5pt}
\noindent\textbf{The Effectiveness of Each Operation in Deliberation Unit}
\setlength{\parskip}{0pt}

\noindent We conduct experiments on DualVD-DAM to reveal the influence of essential operations in the Deliberation Unit: 1) \textbf{I-S} only uses semantic-level image information for information selection. 2) \textbf{I-V} only utilizes visual-level image information for information selection. 3) \textbf{I-SV} jointly exploits semantic and visual information for information selection. 4) \textbf{H} only leverages dialogue history for information selection. 

As shown in Table \ref{abd}, I-S and I-V update information from visual and semantic aspect respectively, while I-SV updates information from both two aspects which achieves the best performance compared to the above two models. The relatively higher results of H model indicate that the history information plays a more important role in the decoder. By jointly incorporating all the structure-aware information from the encoder, DualVD-DAM achieves the best performance on all the metrics. It proves the advantages of DAM via fully utilizing the information from the elaborate encoder, which is beneficial for enhancing the existing generation models by incorporating their encoders with DAM adaptively. 

\begin{table}[t] 
\setlength{\abovecaptionskip}{5pt}
\setlength{\belowcaptionskip}{-4mm} 
\centering
\resizebox{.9\columnwidth}{!}{
\begin{tabular} {L{2.2cm}C{0.6cm}C{0.6cm}C{0.6cm}C{0.69cm}C{0.6cm}C{0.77cm}}
\hline                       
Model & MRR & R\textsl{@}1 & R\textsl{@}5 & R\textsl{@}10 & Mean & NDCG  \\
\hline  
I-S&50.01&40.25&59,78&66.76&17.67&59.09\\
I-V&50.03&40.30&59.34&66.90&17.34&58.93\\
I-SV&50.13&40.34&60.09&67.06&17.34&59.51\\
\hline
H&50.19&40.36&60.09&66.96&17.27&59.92\\
\hline
\textbf{DualVD-DAM} &\textbf{50.51}&\textbf{40.53}&\textbf{60.84}&\textbf{67.94}&\textbf{16.65}&\textbf{60.93 }\\
\hline  
\end{tabular}  
}
\caption{Ablation study of Deliberation Unit on VisDial v1.0.}
\label{abd}
\end{table}

\subsection{Qualitative Analysis}

\paragraph{Response generation quality.} Figure \ref{visual-se} shows two examples of three-round dialogues and the corresponding responses generated by DualVD-G and DualVD-DAM. When answering Q3 in the first example, DualVD-DAM generates accurate and non-repetitive response ``\emph{having fun}" compared with DualVD-G. Comparing to the human response, DualVD-DAM further provides detailed description ``\emph{he is posing for the picture}" so as to increase the richness of the response. Similar observation exists in the second example. 

\paragraph{Information selection quality.} We further visualize the evidence captured from the image and dialogue history for generating the essential words, i.e. \emph{there}, \emph{buildings} and \emph{background}. As shown in Figure \ref{time}, it is difficult to answer the question of ``\emph{Is there a building?}" accurately, since the \emph{buildings} are distant and small. DualVD-DAM accurately focuses on the visual and dialogue clues. Taking the word \emph{background} for example, our model focuses on the background in the image and highlights the \emph{clear day} in the dialogue history. It proves that DAM can adaptively focus on the exact visual and textual clues for generating each word, which contributes to the high quality of the responses.

\section{Conclusion}
In this paper, we propose a novel generative decoder DAM consisting of the {\it Deliberation} Unit,  {\it  Abandon}  Unit and  {\it Memory} Unit.  The novel decoder adopts a compositive decoding mode in order to model information from both response-level and word-level, so as to discourage repetition in the generated responses. DAM is a universal decoding architecture which can be incorporated with existing visual dialogue encoders to improve their performance. The extensive experiments of combining DAM with LF, MN and DualVD encoders verify that our proposed DAM can effectively improve the generation performance of existing models and achieve new state-of-the-art results on the popular benchmark dataset.

\section*{Acknowledgements}
This work is supported by the National Key Research and Development Program (Grant No.2017YFB0803301).

\bibliographystyle{named}
\bibliography{ijcai20}

\end{document}